\documentclass[preprints,article,accept,moreauthors,pdftex]{Definitions/mdpi}
\usepackage{bm}

\usepackage{inconsolata}

\usepackage{graphicx}
\usepackage{multirow}
\usepackage{subcaption}    

\usepackage{booktabs}       
\usepackage{enumitem}       
\usepackage{amssymb}        
\usepackage[hang,flushmargin]{footmisc}  
\usepackage{array}
\usepackage{changepage} 
\usepackage{listings}
\setenumerate[1]{leftmargin=*}    
\setitemize[1]{leftmargin=*}      

\usepackage{xspace}
\newcommand{\EXE}{\texttt{EXE}\xspace}
\newcommand{\ESM}{\texttt{ESM}\xspace}
\newcommand{\ETM}{\texttt{ETM}\xspace}

\newcommand{\JOIN}{\texttt{\textcolor{blue}{JOIN}}\xspace}
\newcommand{\DISTINCT}{\texttt{\textcolor{blue}{DISTINCT}}\xspace}
\newcommand{\NONNULL}{\texttt{\textcolor{blue}{NON\_NULL}}\xspace}
\newcommand{\UNIQUE}{\texttt{\textcolor{blue}{UNIQUE}}\xspace}
\newcommand{\AVE}{\texttt{\textcolor{blue}{AVE}}\xspace}
\newcommand{\COUNT}{\texttt{\textcolor{blue}{COUNT}}\xspace}
\newcommand{\LIMIT}{\texttt{\textcolor{blue}{LIMIT}}\xspace}
\newcommand{\LEFTJOIN}{\textcolor{blue}{\texttt{LEFT} \texttt{JOIN}\xspace}}
\newcommand{\RIGHTJOIN}{\textcolor{blue}{\texttt{RIGHT} \texttt{JOIN}\xspace}}
\newcommand{\INNERJOIN}{\textcolor{blue}{\texttt{INNER} \texttt{JOIN}\xspace}}
\newcommand{\OUTERJOIN}{\textcolor{blue}{\texttt{OUTER} \texttt{JOIN}\xspace}}

\newcommand{\CAST}{\texttt{\textcolor{blue}{CAST}}\xspace}
\newcommand{\CASE}{\texttt{\textcolor{blue}{CASE}}\xspace}

\newcommand{\IN}{\textcolor{blue}{\texttt{IN}}\xspace}
\newcommand{\AS}{\textcolor{blue}{\texttt{AS}}\xspace}

\firstpage{1} 
\pubvolume{1}
\issuenum{1}
\articlenumber{0}
\pubyear{2022}
\copyrightyear{2022}
\datereceived{} 
\dateaccepted{} 
\datepublished{} 
\hreflink{https://doi.org/} 

\Title{ETM: Modern Insights into Perspective on Text-to-SQL Evaluation in the Age of Large Language Models}

\TitleCitation{ETM: Modern Insights into Perspective on Text-to-SQL Evaluation in the Age of Large Language Models}

\Author{Benjamin Ascoli $^{1\dagger*}$, Yasoda Sai Ram Kandikonda $^{2\dagger}$ and Jinho D. Choi $^{3\dagger*}$}

\AuthorNames{Benjamin Ascoli, Yasoda Sai Ram Kandikonda and Jinho Choi}

\AuthorCitation{Ascoli, B.; Kandikonda, Y.; Choi, J.}

\address{%

$^{1}$ \quad bascoli@emory.edu (B.A.)\\
$^{2}$ \quad ykandik@emory.edu (Y.K.)\\
$^{3}$ \quad jinho.choi@emory.edu (J.C.)}

\corres{Correspondence: bascoli@emory.edu (B.A.); jinho.choi@emory.edu (J.C.)}

\firstnote{Current address: Department of Computer Science, Emory University, Atlanta, GA, USA}

\abstract{The task of Text-to-SQL enables anyone to retrieve information from SQL databases using natural language.
While this task has made substantial progress, the two primary evaluation metrics--Execution Accuracy (\EXE) and Exact Set Matching Accuracy (\ESM)--suffer from inherent limitations that can misrepresent performance.
Specifically, \ESM’s rigid matching overlooks semantically correct but stylistically different queries, whereas \EXE can overestimate correctness by ignoring structural errors that yield correct outputs.
These shortcomings become especially problematic when assessing outputs from large language model (LLM)-based approaches without fine-tuning, which vary more in style and structure compared to their fine-tuned counterparts.
Thus, we introduce a new metric, Enhanced Tree Matching (\ETM), which mitigates these issues by comparing queries using both syntactic and semantic elements.
Through evaluating nine LLM-based models, we show that \EXE and \ESM can produce false positive and negative rates as high as 23.0\% and 28.9\%, while \ETM reduces these rates to 0.3\% and 2.7\%, respectively. 
We release our \ETM script as open source, offering the community a more robust and reliable approach to evaluating Text-to-SQL.}

\keyword{text-to-sql; evaluation metrics; large language models} 

\begin{document}



\section{Introduction}
\label{sec:introduction}

While interacting with SQL databases through natural language interfaces makes them significantly more accessible to non-experts, the task of automatically mapping natural language requests to SQL queries for relational databases, known as Text-to-SQL, remains challenging.
Lately, the advent of the transformer \cite{vaswani2017attention} and large language models (LLMs) \cite{brown2020languageGPT3,raffel:20} has led to momentous advancements in this field.
Notably, LLMs have overcome several challenges in Text-to-SQL, evidenced by their dominance in leaderboards for popular benchmarks like the Spider dataset \cite{yu2019spider} and the more challenging BIRD dataset \citep{BIRD2023}, underscoring their effectiveness in handling complex, multi-table SQL query generation that previous approaches had struggled with.

Evaluating Text-to-SQL models is also challenging because SQL equivalence has been shown to be undecidable \cite{foundationsofdatabase1995}.
Text-to-SQL models are tested using two metrics: Execution Accuracy (\textbf{\EXE}) and Exact Set Matching Accuracy (\textbf{\ESM}).
\EXE checks if the SQL execution result of the predicted query matches that of the gold standard query.
However, \EXE can yield false positives, as semantically different queries may produce the same output (Figure~\ref{fig:EXFalsePositive}).
On the other hand, \ESM assesses the predicted query by comparing sets of keywords and their arguments to those of the gold query.
While more rigorous than \EXE, \ESM is prone to false negatives, because SQL queries may be semantically equivalent yet syntactically diverse (Figure~\ref{fig:EMFalseNegative}).
These issues raise the need for a more robust evaluation metric that accurately evaluates the performance of Text-to-SQL models.

\begin{figure}[htbp!]
\centering

\begin{subfigure}{\columnwidth}
\centering

\begin{minipage}{.9\textwidth}
\begin{lstlisting}[language=SQL,keywordstyle=\color{blue},numbers=none,framexleftmargin=0.2em,xleftmargin=0.4em]
SELECT name FROM dogs;
\end{lstlisting}

\begin{lstlisting}[language=SQL,keywordstyle=\color{blue},numbers=none,framexleftmargin=0.2em,xleftmargin=0.4em]
SELECT name FROM dogs WHERE age < 10;
\end{lstlisting}
\end{minipage}

\caption{Semantically distinct queries having the same execution result, as there are no dogs with age $\geq$ 10.}
\label{fig:EXFalsePositive}
\end{subfigure}
\vspace{-0.5em}

\begin{subfigure}{\columnwidth}
\centering

\begin{minipage}{0.9\textwidth}
\begin{lstlisting}[language=SQL,keywordstyle=\color{blue},numbers=none,framexleftmargin=0.2em,xleftmargin=0.4em]
SELECT MAX(weight) FROM dogs;
\end{lstlisting}

\begin{lstlisting}[language=SQL,keywordstyle=\color{blue},numbers=none,framexleftmargin=0.2em,xleftmargin=0.4em]
SELECT weight FROM dogs ORDER BY weight DESC LIMIT 1;
\end{lstlisting}
\end{minipage}

\caption{Syntactically distinct but semantically equivalent queries to find the weight of the heaviest dog.}
\label{fig:EMFalseNegative}
\end{subfigure}
\vspace{-1em}
\caption{Examples of a false positive yielded by \EXE (\ref{fig:EXFalsePositive}) and a false negative yielded by \ESM (\ref{fig:EMFalseNegative}).}
\label{fig:false-pn-examples}
\end{figure}

Models using pretrained LLMs without fine-tuning, such as GPT (henceforth \textbf{PLM}), perform particularly well on \EXE, which is the main metric used on the Spider and BIRD leaderboards.
Surprisingly, they do not show a similar level of performance on \ESM.
When using \ESM as the primary metric, no PLM-based models rank highly, a stark contrast to the Spider and BIRD leaderboards.
Therefore, it is critical to examine these metrics and determine the most suitable approach for an accurate evaluation of model performance, as the disparity between them disproportionately impacts PLM-based models compared to those using fine-tuned LLMs (henceforth, \textbf{FLM}).

This paper first examines potential issues in \EXE and \ESM, and proposes a new enhanced metric, called \textbf{\ETM}, which addresses many shortcomings present in the original metrics (Section~\ref{sec:revision}). 
Nine state-of-the-art models are evaluated on the Spider and BIRD datasets, comparing their performance using \EXE, \ESM, and \ETM (Section~\ref{sec:evaluation_existing}). 
Finally, a comprehensive error analysis is conducted on the evaluation results using these three metrics, revealing the superior robustness of \ETM (Section~\ref{sec:analysis}).
We posit that \ETM will serve as a pivotal metric for assessing the real capabilities of LLM-based Text-to-SQL models, thereby enabling them to reach new heights of performance.\footnote{All our resources, including the new evaluation script and the model outputs, are available through our open-source project: \url{https://github.com/emorynlp/ETM}}

\section{Related Work}
\label{sec:related-work}
\subsection{Text-to-SQL Models}

The current state-of-the-art performance has been achieved by PLM-based models using GPT 
\cite{openai-2024-gpt4}.
\citet{dong2023c3} introduced C3, which employs schema linking to rank relevant tables and columns and prompts GPT to generate SQL queries.
\citet{pourreza2023dinsql} proposed DIN-SQL, which predicts schema links, classifies query difficulty, and prompts GPT using template-based queries with debugging prompts.
\citet{gao2023texttosql} presented DAIL-SQL, which searches for similar questions in the training set and uses them to create a few-shot prompt with GPT to generate an initial query.
This is then used to find more similar queries in the training set, and the most similar ones are used in a second few-shot prompt to generate the final query.
Despite achieving high ranks on the Spider leaderboard, evaluated on \EXE, none of these PLM-based models appear on the CoSQL leaderboard, evaluated on \ESM \cite{yu2019cosql}.

Several FLM-based models, such as fine-tuned T5 \cite{raffel:20}, have also been introduced, showing comparable results to PLM-based models on Spider.
RASAT \cite{qi2022rasat}, coupled with PICARD \cite{scholak2021picard}, incorporates relation-aware self-attention, enabling better schema understanding while inheriting pre-trained weights from T5.
\citet{li2023graphixt5} introduced Graphix-T5, which augments T5 with graph-aware layers to integrate semantic information from transformer blocks with structural information from graph neural networks.
\citet{li2023resdsql} presented RESDSQL that utilizes an encoder to identify relevant schema items and a decoder to first generate the SQL skeleton with keywords.
\citet{li2024codesbuildingopensourcelanguage} introduced CodeS, an open-source series of language models specifically designed for text-to-SQL. CodeS undergoes incremental pre-training with a curated SQL-centric corpus, and uses schema filtering and prompt construction techniques.
SuperSQL \cite{supersql2024}, a hybrid FLM/PLM framework, uses a genetic learning algorithm to swap model components to improve its output.

\subsection{Evaluation of SQL Equivalence}
\label{ssec:sql-equivalence}

Although evaluating the equivalence of two queries plays a crucial role in advancing Text-to-SQL models, few works have addressed this challenge.
\citet{Cosette2017} introduced Cosette, an automatic SQL solver that compiles queries over relational tables and checks their semantic equivalence, producing counterexamples when not equivalent; however, it supports limited SQL operations. 
\citet{DBLP:journals/pvldb/ZhouANHX19EQUITAS} presented EQUITAS, an automated verification tool that transforms SQL queries into first-order logic to verify equivalence, although its source code is not publicly available.
\citet{zhong2020semantic} proposed test-suite execution matching to measure semantic accuracy, which generates a small suite of slightly altered databases to help reduce the false positive rate of \EXE.
However, this approach is not scalable and suffers from the long execution times of some queries, especially when dealing with larger databases.
More recently, \citet{nooralahzadeh-etal-2024-statbotswiss} introduced soft and partial execution accuracy, which aimed to reduce error from ambiguous questions by allowing multiple answers to be correct. 
However, in doing this, it relaxes the definition of semantic equivalence, allowing more false positives. 
\citet{tsed2024} evaluated SQL queries using the editing difference between their abstract syntax trees (TSED), which faces challenges as two queries can vary in structure but still be equivalent.
\citet{zhan2025-funcevalgmn} proposed FuncEvalGMN, which compares queries by transforming them into relational operator trees and using a graph matching network to assess functional equivalence by comparing their embeddings. 
However, it relies on an embedding similarity threshold for accurate results, which can struggle with subtle semantic differences and threshold tuning.

Therefore, the most accessible and widely used automatic evaluation approaches for Text-to-SQL remain \EXE and \ESM \citep{yu2019spider}.
Their combined evaluation script provides options to disable value and distinct checks, which were employed due to prior model limitations.
However, despite the proficiency of LLM-based models in handling these aspects, the results in the literature for Spider are still reported with both value and distinct checking disabled, obscuring the true performance of LLM-based models in real-world applications.
BIRD uses \EXE as its primary metric, but evaluates the query output without regard to duplicate rows or ordering.
Since this can lead to even more false positives, we focus primarily on the original \EXE metric, which has more strict output matching.
\section{Materials and Methods}
\label{sec:revision}

For a comprehensive analysis of the two metrics, Execution Accuracy (\EXE) and Exact Set Matching (\ESM), we evaluate nine models on the Spider \cite{yu2019spider} and BIRD \cite{BIRD2023} datasets. 
Cases of false positives (Section~\ref{flaws}) and negatives (Section~\ref{fixes}) in \ESM are thoroughly examined through this analysis, and addressed in our new metric, \ETM (Section~\ref{ssec:etm}).

\subsection{False Positives in ESM}
\label{flaws}

We first analyze the queries predicted by the models along with their gold standard counterparts that are considered equivalent by \ESM but not by \EXE.
Since \ESM is a more stringent metric, it is expected that no query pair considered a mismatch by \EXE would be considered a match by \ESM.
Upon closer inspection, however, it becomes evident that \ESM has several shortcomings in its evaluation approach.

\lstset{
  escapeinside={(*@}{@*)}
}
\begin{figure}[htbp]
\centering
\begin{minipage}{.9\columnwidth}
\begin{lstlisting}[language=SQL,keywordstyle=\color{blue},numbers=none,framexleftmargin=0.2em,xleftmargin=0.4em]
SELECT * FROM dogs AS t1 JOIN breeds AS t2 ON t1.breed_code = (*@\textbf{t2.breed\_code}@*);
\end{lstlisting}

\begin{lstlisting}[language=SQL,keywordstyle=\color{blue},numbers=none,framexleftmargin=0.2em,xleftmargin=0.4em]
SELECT * FROM dogs AS t1 JOIN breeds AS t2 ON t1.breed_code = (*@\textbf{t2.breed\_name}@*);
\end{lstlisting}
\end{minipage}
\caption{A query pair, correctly considered a mismatch by \EXE, but considered a match by \ESM.}
\label{joinexactfail}
\end{figure}

\noindent One major issue is that \ESM  does not account for \JOIN conditions, which are essential parts of many SQL queries.
In Figure~\ref{joinexactfail}, the two queries produce different outputs such that \EXE \textit{correctly} considers them a mismatch.
\ESM \textit{mistakenly} considers them a match, however, because it ignores the \JOIN conditions (\texttt{t2.breed\_code} vs. \texttt{t2.breed\_name}). 

\begin{figure}[htbp]
\centering
\begin{minipage}{.9\columnwidth}
\begin{lstlisting}[language=SQL,keywordstyle=\color{blue},numbers=none,framexleftmargin=0.2em,xleftmargin=0.4em]
SELECT DISTINCT name FROM dogs;
\end{lstlisting}

\begin{lstlisting}[language=SQL,keywordstyle=\color{blue},numbers=none,framexleftmargin=0.2em,xleftmargin=0.4em]
SELECT name FROM dogs;
\end{lstlisting}
\end{minipage}
\caption{A query pair mistakenly considered a match by \ESM,  as it overlooks the \DISTINCT keyword.}
\label{distinctexactfail}
\end{figure}

\noindent Another issue arises when evaluating queries with the \DISTINCT keyword.
Even when distinct checks are enabled in the \ESM script (Section~\ref{ssec:sql-equivalence}), it considers \DISTINCT only within aggregate keywords, like \COUNT or \AVE, failing to recognize it in simpler and more commonly used cases (Figure~\ref{distinctexactfail}).

\begin{figure}[htbp]
\centering
\begin{minipage}{.9\columnwidth}
\begin{lstlisting}[language=SQL,keywordstyle=\color{blue},numbers=none,framexleftmargin=0.2em,xleftmargin=0.4em]
SELECT transcript_date FROM Transcripts ORDER BY transcript_date DESC LIMIT (*@\textbf{2}@*);
\end{lstlisting}

\begin{lstlisting}[language=SQL,keywordstyle=\color{blue},numbers=none,framexleftmargin=0.2em,xleftmargin=0.4em]
SELECT transcript_date FROM Transcripts ORDER BY transcript_date DESC LIMIT (*@\textbf{1}@*);
\end{lstlisting}
\end{minipage}
\vspace{-0.5em}
\caption{A query pair mistakenly considered a match by \ESM due to its disregard of the \LIMIT values.}
\label{limitvaluefail}
\end{figure}

\noindent Additionally, the \ESM script ignores specified \LIMIT values even when value checks are enabled (Figure~\ref{limitvaluefail}).

\subsection{False Negatives in ESM}
\label{fixes}

We also analyze the predicted and gold query pairs that \EXE finds equivalent but not \ESM.
Some of these cases are false positives for \EXE, where the queries are semantically distinct but still return the same result when executed. The other cases involve queries that are semantically equivalent but syntactically distinct, causing \ESM to mistakenly consider them a mismatch. 
These false negatives occur because assessing semantic equivalence is often contingent on certain assumptions about the database.

\begin{figure}[htb!]
\centering
\begin{minipage}{.9\columnwidth}
\begin{lstlisting}[language=SQL,keywordstyle=\color{blue},numbers=none,framexleftmargin=0.2em,xleftmargin=0.4em]
SELECT count(dog_id) FROM dogs;
\end{lstlisting}

\begin{lstlisting}[language=SQL,keywordstyle=\color{blue},numbers=none,framexleftmargin=0.2em,xleftmargin=0.4em]
SELECT count(*) FROM dogs;
\end{lstlisting}
\end{minipage}
\vspace{-0.5em}
\caption{A semantically equivalent query pair under a verifiable assumption (\texttt{dog\_id} is \NONNULL).}
\label{countpk}
\end{figure}

\noindent In Figure~\ref{countpk}, the queries are semantically equivalent only if the column \texttt{dog\_id} is \NONNULL.
This can be verified by the database schema, which gives information about tables and columns, such as primary-foreign key relationships and constraints.
Likewise, the queries in Figure~\ref{distinctexactfail} can also be considered a match if the column \texttt{name} in the table \texttt{dogs} is \UNIQUE.
To this end, we carefully examine every false negative case and compile verifiable assumptions that are sufficiently general for any database schema to alleviate this challenge.

\subsection{New Evaluation Metric}
\label{ssec:etm}

\begin{figure}[htb!]
\centering
\includegraphics[width=0.9\columnwidth]{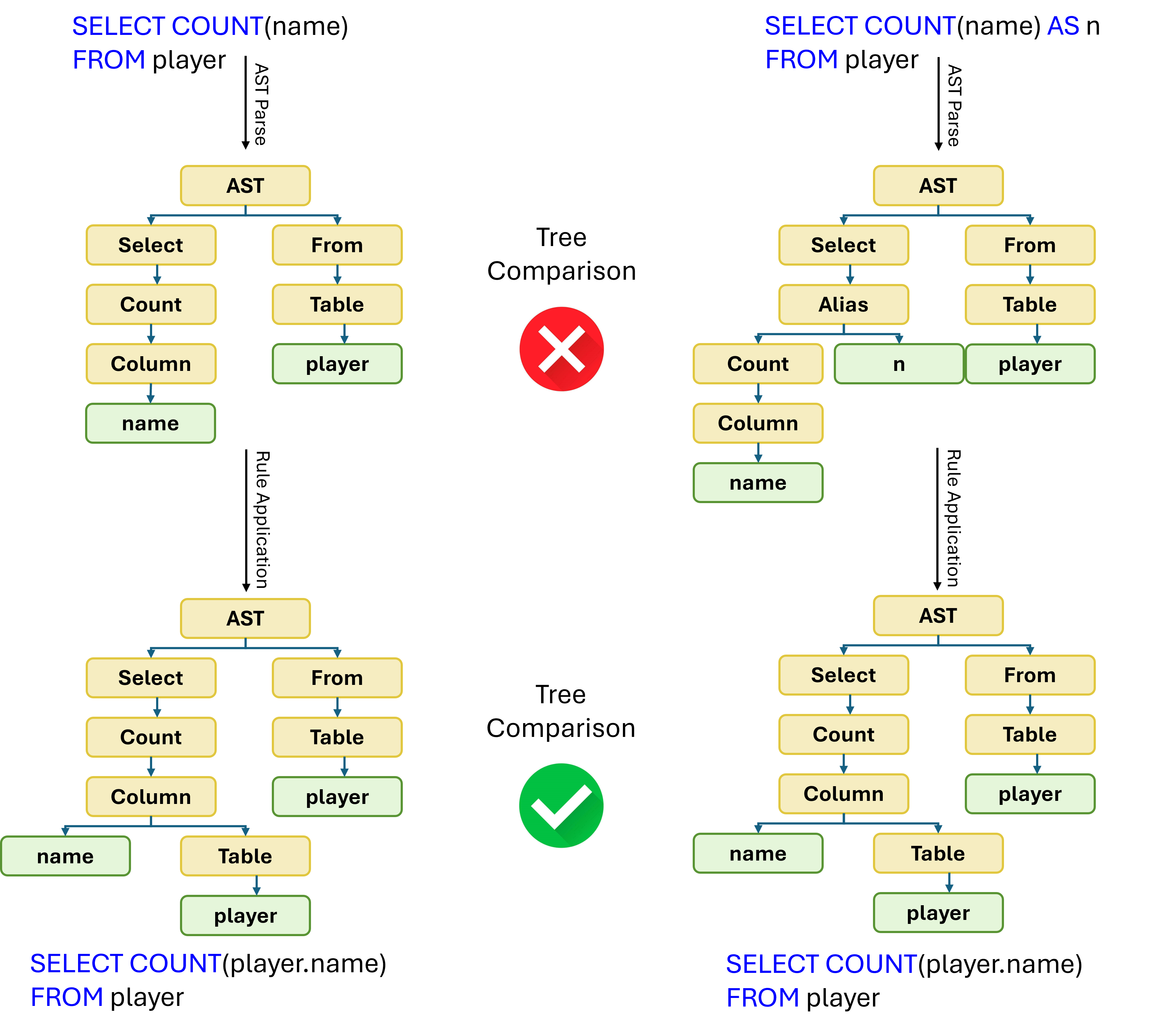}
\caption{Overview of the Enhanced Tree Matching (\ETM) process. Queries are parsed into their ASTs, which are then normalized using a set of predefined equivalence rules (Table~\ref{tab:rules_assumptions}) before comparison.}
\label{fig:etm_pipeline}
\end{figure}

We present Enhanced Tree Matching (\ETM), a new evaluation metric that compares queries based on their abstract syntax tree (AST), rather than the set-based matching approach of \ESM. 
\ETM applies a set of verifiable equivalence rules to transform queries into normalized forms before comparison, reducing false negatives present in \ESM (Section~\ref{fixes}).
Figure~\ref{fig:etm_pipeline} shows a high-level overview of \ETM's process for comparing SQL queries.
Queries are parsed into Abstract Syntax Trees (ASTs), then normalized using predefined equivalence rules such that the normalization recognizes that structural differences like aliases do npt affect query meaning.
After normalization, the originally different ASTs become equivalent, enabling semantic rather than textual query comparison.

\noindent \ETM addresses all the issues in Sections \ref{flaws} and \ref{fixes}, as well as other critical issues.
Tables~\ref{tab:p_rules_assumptions}, \ref{tab:rules_assumptions}, and \ref{tab:rules_no_assumptions} provide a full list of equivalent queries and verifiable assumptions incorporated into \ETM.
The preprocessing rules (Table~\ref{tab:p_rules_assumptions}) handle basic syntactic variations like case sensitivity, table prefixes, column ordering, and aliasing, while the advanced equivalence rules (Tables~\ref{tab:rules_assumptions} and \ref{tab:rules_no_assumptions}) capture more complex semantic relationships that require database schema constraints for verification.

\lstset{
    basicstyle=\small\ttfamily,
    columns=fullflexible,
    breaklines=true,
    breakatwhitespace=true,
    postbreak=\mbox{\textcolor{red}{$\hookrightarrow$}\space},
    breakindent=0pt,
    captionpos=b,
    numbers=left,  
    numberstyle=\tiny\color{gray},  
    stepnumber=1,  
    numbersep=4pt,  
    xleftmargin=0em,  
    xrightmargin=0em,  
    frame=single,  
    framexleftmargin=0em,  
    aboveskip=2mm,
    belowskip=0mm,
}

\begin{table}[htpb!]
\centering\small
\caption{Preprocessing equivalence rules implemented in \ETM. \textbf{\texttt{t*}}: table, \textbf{\texttt{c*}}: column, \textbf{\texttt{d*}}: condition.}
\label{tab:p_rules_assumptions}

\begin{tabular}{c|l|c}

\textbf{ID} & \multicolumn{1}{c|}{\textbf{Equivalent Queries}} & \textbf{Verifiable Assumptions} \\
\hline
\multirow{2}{*}{P0} &
\begin{lstlisting}[language=SQL,keywordstyle=\color{blue},numbers=none,frame=none]
SELECT c1 FROM t1;
\end{lstlisting}\vspace{-0.5em}
& \multirow{2}{*}{None} \\ 
& \begin{lstlisting}[language=SQL,keywordstyle=\color{blue},numbers=none,frame=none]
SELECT C1 FROM T1;
\end{lstlisting}
& \\
\hline
\multirow{2}{*}{P1} &
\begin{lstlisting}[language=SQL,keywordstyle=\color{blue},numbers=none,frame=none]
SELECT c1 FROM t1;
\end{lstlisting}\vspace{-0.5em}
& \multirow{2}{*}{None} \\ 
&  \begin{lstlisting}[language=SQL,keywordstyle=\color{blue},numbers=none,frame=none]
SELECT t1.c1 FROM t1;
\end{lstlisting}
& \\
\hline
\multirow{2}{*}{P2} &
 \begin{lstlisting}[language=SQL,keywordstyle=\color{blue},numbers=none,frame=none]
SELECT c1, c2 FROM t1;
\end{lstlisting}\vspace{-0.5em}
& \multirow{2}{*}{None} \\ 
&  \begin{lstlisting}[language=SQL,keywordstyle=\color{blue},numbers=none,frame=none]
SELECT c2, c1 FROM t1;
\end{lstlisting}
& \\
\hline
\multirow{2}{*}{P3} &
 \begin{lstlisting}[language=SQL,keywordstyle=\color{blue},numbers=none,frame=none]
SELECT t1.c1 FROM t1;
\end{lstlisting} \vspace{-0.5em}
& \multirow{2}{*}{None} \\ 
&  \begin{lstlisting}[language=SQL,keywordstyle=\color{blue},numbers=none,frame=none]
SELECT t.c1 FROM t1 AS t;
\end{lstlisting}
& \\
\hline
\multirow{2}{*}{P4} &
 \begin{lstlisting}[language=SQL,keywordstyle=\color{blue},numbers=none,frame=none]
SELECT _ FROM t1 JOIN t2;
\end{lstlisting} \vspace{-0.5em}
& \multirow{2}{*}{None} \\ 
& \begin{lstlisting}[language=SQL,keywordstyle=\color{blue},numbers=none,frame=none]
SELECT _ FROM t2 JOIN t1;
\end{lstlisting}
& \\
\hline
\multirow{2}{*}{P5} &
\begin{lstlisting}[language=SQL,keywordstyle=\color{blue},numbers=none,frame=none]
SELECT _ FROM _ WHERE x =/AND/OR y;
\end{lstlisting} \vspace{-0.5em}
& \multirow{2}{*}{None} \\ 
&  \begin{lstlisting}[language=SQL,keywordstyle=\color{blue},numbers=none,frame=none]
SELECT _ FROM _ WHERE y =/AND/OR x;
\end{lstlisting}
& \\
\hline
\multirow{2}{*}{P6} &
\begin{lstlisting}[language=SQL,keywordstyle=\color{blue},numbers=none,frame=none]
SELECT col AS c FROM t1;
\end{lstlisting} \vspace{-0.5em}
& \multirow{2}{*}{None} \\ 
&  \begin{lstlisting}[language=SQL,keywordstyle=\color{blue},numbers=none,frame=none]
SELECT col FROM t1;
\end{lstlisting}

& \\
\hline
\multirow{2}{*}{P7} &
 \begin{lstlisting}[language=SQL,keywordstyle=\color{blue},numbers=none,frame=none]
SELECT _ FROM _ WHERE d1;
\end{lstlisting}\vspace{-0.5em}
& \multirow{2}{*}{None} \\ 
& \begin{lstlisting}[language=SQL,keywordstyle=\color{blue},numbers=none,frame=none]
SELECT _ FROM _ WHERE (d1);
\end{lstlisting}
& \\
\hline
\multirow{2}{*}{P8} &
 \begin{lstlisting}[language=SQL,keywordstyle=\color{blue},numbers=none,frame=none]
SELECT "t1"."c1" FROM "t1";
\end{lstlisting}\vspace{-0.5em}
& \multirow{2}{*}{None} \\ 
&  \begin{lstlisting}[language=SQL,keywordstyle=\color{blue},numbers=none,frame=none]
SELECT t1.c1 FROM t1;
\end{lstlisting}
& 
\end{tabular}
\vspace{-1em}
\end{table}

\lstset{
  aboveskip=-2mm,
  belowskip=-4mm,
  basicstyle=\ttfamily\small,
  numbers=none,
  frame=none,
  language=SQL
}
\begin{table}[htp!]
\begin{adjustwidth}{-\extralength}{0cm}
\centering\small
\caption{Equivalent queries with verifiable assumptions implemented in \ETM. \textbf{\texttt{t*}}: table, \textbf{\texttt{c*}}: column, \textbf{\texttt{d*}}: condition, \textbf{\texttt{q*}}: full query. \textbf{Case 1}: a primary key-foreign key relation, where \texttt{t1.c1} is the primary key and \texttt{t2.c2} is the foreign key. \textbf{Case 2}: \texttt{t1.c1} must be non-composite and \texttt{X} can be any column(s) in \texttt{t2}. \textbf{\texttt{/}} denotes options, but consistency is required in selecting between options across corresponding elements of the queries.}
\label{tab:rules_assumptions}

\begin{tabular}{c|>{\raggedright\arraybackslash}p{13cm}|c}

\textbf{ID} & \multicolumn{1}{c|}{\textbf{Equivalent Queries}} & \textbf{Verifiable Assumptions} \\
\hline
\multirow{2}{*}{1} &
\begin{lstlisting}[language=SQL,keywordstyle=\color{blue},numbers=none,frame=none]
SELECT _ FROM t1 WHERE c1 = (SELECT MIN/MAX(c1) FROM t1);
\end{lstlisting}\vspace{-0.5em}
& \multirow{2}{*}{\textbf{\texttt{c1}} is \UNIQUE} \\
& \begin{lstlisting}[language=SQL,keywordstyle=\color{blue},numbers=none,frame=none]
SELECT _ FROM t1 ORDER BY c1 ASC/DESC LIMIT 1;
\end{lstlisting} 
& \\
\hline

\multirow{2}{*}{2} &
 \begin{lstlisting}[language=SQL,keywordstyle=\color{blue},numbers=none,frame=none]
SELECT DISTINCT c1 FROM t1;
\end{lstlisting}\vspace{-0.5em} 
& \multirow{2}{*}{\textbf{\texttt{c1}} is \UNIQUE} \\
& \begin{lstlisting}[language=SQL,keywordstyle=\color{blue},numbers=none,frame=none]
SELECT c1 FROM t1;
\end{lstlisting} 
& \\
\hline

\multirow{2}{*}{3} &
 \begin{lstlisting}[language=SQL,keywordstyle=\color{blue},numbers=none,frame=none]
SELECT c1 FROM t1 WHERE d1 INTERSECT/UNION SELECT c1 FROM t1 WHERE d2;
\end{lstlisting} \vspace{-0.5em} 
& \multirow{2}{*}{\textbf{\texttt{c1}} is \UNIQUE} \\
&  \begin{lstlisting}[language=SQL,keywordstyle=\color{blue},numbers=none,frame=none]
SELECT c1 FROM t1 WHERE d1 AND/OR d2;
\end{lstlisting} 
& \\
\hline

\multirow{2}{*}{4} &
 \begin{lstlisting}[language=SQL,keywordstyle=\color{blue},numbers=none,frame=none]
SELECT _ FROM t1 WHERE GROUP BY c1,c2,...;
\end{lstlisting} \vspace{-0.5em} 
& \multirow{2}{*}{\textbf{\texttt{c1}} is \UNIQUE} \\
&  \begin{lstlisting}[language=SQL,keywordstyle=\color{blue},numbers=none,frame=none]
SELECT _ FROM t1 WHERE GROUP BY c1;
\end{lstlisting} 
& \\
\hline

\multirow{2}{*}{5} &
 \begin{lstlisting}[language=SQL,keywordstyle=\color{blue},numbers=none,frame=none]
SELECT c1 FROM t1 EXCEPT (q1);
\end{lstlisting} \vspace{-0.5em} 
& \textbf{\texttt{c1}} is \UNIQUE \\
& \begin{lstlisting}[language=SQL,keywordstyle=\color{blue},numbers=none,frame=none]
SELECT c1 FROM t1 WHERE c1 NOT IN (q1);
\end{lstlisting} 
& and \NONNULL \\
\hline

\multirow{2}{*}{6} &
\begin{lstlisting}[language=SQL,keywordstyle=\color{blue},numbers=none,frame=none]
SELECT COUNT(*) FROM t1; 
\end{lstlisting} \vspace{-0.5em}
& \multirow{2}{*}{\textbf{\texttt{c1}} is \NONNULL} \\
&  \begin{lstlisting}[language=SQL,keywordstyle=\color{blue},numbers=none,frame=none]
SELECT COUNT(c1) FROM t1;
\end{lstlisting}
& \\
\hline

\multirow{2}{*}{7} &
\begin{lstlisting}[language=SQL,keywordstyle=\color{blue},numbers=none,frame=none]
SELECT _ FROM t1 WHERE c1 is NOT NULL;
\end{lstlisting} \vspace{-0.5em}
& \multirow{2}{*}{\textbf{\texttt{c1}} is \NONNULL} \\
& \begin{lstlisting}[language=SQL,keywordstyle=\color{blue},numbers=none,frame=none]
SELECT _ FROM t1; 
\end{lstlisting}
& \\
\hline
\multirow{2}{*}{8} &
\begin{lstlisting}[language=SQL,keywordstyle=\color{blue},numbers=none,frame=none]
SELECT CAST(SUM(c1) AS FLOAT) / COUNT(*) FROM t1;
\end{lstlisting}\vspace{-0.5em}
& \multirow{2}{*}{\textbf{\texttt{c1}} is \NONNULL} \\ 
& \begin{lstlisting}[language=SQL,keywordstyle=\color{blue},numbers=none,frame=none]
SELECT AVG(c1) FROM t1;
\end{lstlisting}
& \\
\hline
\multirow{2}{*}{9} &
\begin{lstlisting}[language=SQL,keywordstyle=\color{blue},numbers=none,frame=none]
SELECT COUNT(CASE WHEN d1 THEN 1/c1 ELSE NULL END) FROM t1;
\end{lstlisting}\vspace{-0.5em}
& \multirow{2}{*}{\textbf{\texttt{c1}} is \NONNULL} \\ 
& \begin{lstlisting}[language=SQL,keywordstyle=\color{blue},numbers=none,frame=none]
SELECT SUM(CASE WHEN d1 THEN 1 ELSE 0 END) FROM t1;
\end{lstlisting}
& \\
\hline

\multirow{2}{*}{10} &
\begin{lstlisting}[language=SQL,keywordstyle=\color{blue},numbers=none,frame=none]
SELECT MIN/MAX(c1), _ FROM t1;
\end{lstlisting}\vspace{-0.5em}
& \multirow{2}{*}{\textbf{\texttt{t1}} is not empty} \\
& \begin{lstlisting}[language=SQL,keywordstyle=\color{blue},numbers=none,frame=none]
SELECT c1, _ FROM t1 ORDER BY c1 ASC/DESC LIMIT 1; 
\end{lstlisting}
& \\
\hline

\multirow{2}{*}{11} &
\begin{lstlisting}[language=SQL,keywordstyle=\color{blue},numbers=none,frame=none]
SELECT * FROM t1;
\end{lstlisting}\vspace{-0.5em} 
& \textbf{\texttt{t1}} consists of \\
& \begin{lstlisting}[language=SQL,keywordstyle=\color{blue},numbers=none,frame=none]
SELECT  c1, c2, ... FROM t1;
\end{lstlisting} 
& only \textbf{\texttt{c1}}, \textbf{\texttt{c2}}, {...} \\
\hline

\multirow{2}{*}{12} &
\begin{lstlisting}[language=SQL,keywordstyle=\color{blue},numbers=none,frame=none]
SELECT _ FROM _ WHERE c1 = 'x';
\end{lstlisting}\vspace{-0.5em}
& \textbf{\texttt{x}} is a number not \\ 
& \begin{lstlisting}[language=SQL,keywordstyle=\color{blue},numbers=none,frame=none]
SELECT _ FROM _ WHERE c1 = x; 
\end{lstlisting}
& starting with zero \\
\hline

\multirow{2}{*}{13} &
\begin{lstlisting}[language=SQL,keywordstyle=\color{blue},numbers=none,frame=none]
SELECT _ FROM t2 WHERE c2 IN (SELECT c1 FROM t1 WHERE d1);
\end{lstlisting}\vspace{-0.5em}
& \textbf{Case 1} \\ 
& \begin{lstlisting}[language=SQL,keywordstyle=\color{blue},numbers=none,frame=none]
SELECT _ FROM t1 JOIN t2 ON t1.c1 = t2.c2 WHERE d1; 
\end{lstlisting}
& (refer to the caption) \\ 
\hline

\multirow{2}{*}{14} &
\begin{lstlisting}[language=SQL,keywordstyle=\color{blue},numbers=none,frame=none]
SELECT X FROM t1 JOIN t2 on t1.c1 = t2.c2;
\end{lstlisting}\vspace{-0.5em} 
& \textbf{Case 2} \\
& \begin{lstlisting}[language=SQL,keywordstyle=\color{blue},numbers=none,frame=none]
SELECT X from t2;
\end{lstlisting} 
& (refer to the caption) \\ 
\hline
\multirow{2}{*}{15} &
\begin{lstlisting}[language=SQL,keywordstyle=\color{blue},numbers=none,frame=none]
SELECT _ FROM _ WHERE SUBSTR(c1, 1, a) = x AND SUBSTR(c1, b, c) >/</>=/<= y;
\end{lstlisting}\vspace{-0.5em}
& \multirow{2}{*}{\textbf{\texttt{a}} + 1 = \textbf{\texttt{b}}} \\ 
& \begin{lstlisting}[language=SQL,keywordstyle=\color{blue},numbers=none,frame=none]
SELECT _ FROM _ WHERE c1 >/</>=/<= xy;
\end{lstlisting}
& \\
\hline
\multirow{2}{*}{16} &
\begin{lstlisting}[language=SQL,keywordstyle=\color{blue},numbers=none,frame=none]
SELECT _ FROM _ WHERE c1 LIKE 'x%';
\end{lstlisting}\vspace{-0.5em}
& \multirow{2}{*}{len(\textbf{\texttt{x}}) = \textbf{\texttt{n}}} \\ 
& \begin{lstlisting}[language=SQL,keywordstyle=\color{blue},numbers=none,frame=none]
SELECT _ FROM _ WHERE SUBSTR(c1, 1, n) = 'x'
\end{lstlisting}
& 
\end{tabular}
\end{adjustwidth}
\end{table}
\lstset{
    basicstyle=\footnotesize\ttfamily,
    columns=fullflexible,
    breaklines=true,
    breakatwhitespace=true,
    postbreak=\mbox{\textcolor{red}{$\hookrightarrow$}\space},
    breakindent=0pt,
    captionpos=b,
    numbers=left,  
    numberstyle=\tiny\color{gray},  
    stepnumber=1,  
    numbersep=4pt,  
    xleftmargin=1.2em,  
    xrightmargin=-1.2em,  
    frame=single,  
    framexleftmargin=1em,  
    keepspaces=true,
    aboveskip=2mm,
    belowskip=0mm,
}

\noindent The following details additional key updates in \ETM.


\begin{enumerate}
\item \textbf{Keywords:} The keywords \LEFTJOIN, \RIGHTJOIN, \OUTERJOIN, \INNERJOIN, \CAST, \CASE, and others, previously disregarded by \ESM, are now properly considered.

\item \textbf{Foreign Key Preservation:} \ESM rebuilds queries such that all foreign keys become their primary key counterparts, causing false positives. 
In \ETM, all foreign keys are preserved.

\item \textbf{Join Conditions:} \ESM never compares \JOIN conditions between queries. 
Conditions for any \JOIN are correctly assessed by \ETM. 

\item \textbf{Local Aliases:} \ESM extends aliases to the entire query, causing issues in subqueries where aliases are local. 
\ETM properly scopes aliases to their corresponding subqueries (Listing~\ref{lst:global_alias}).

\vspace{0.5em}
\begin{lstlisting}[language=SQL,keywordstyle=\color{blue},numbers=none,framexleftmargin=0.3em,xleftmargin=.7em,xrightmargin=.3em,caption=\ESM evaluates this incorrectly as it does not recognize that t is not only an alias for t3 in the subquery but also for t1 in the outer query., label=lst:global_alias]
SELECT c1 FROM t1 AS t JOIN t2 ON t.c1=t2.c2 WHERE c1 IN (SELECT c3 FROM t3 AS t);
\end{lstlisting}
\vspace{0.5em}

\item \textbf{DISTINCT:} While \ESM checks for \DISTINCT only within aggregate functions, \ETM consistently considers it across the entire query (Section~\ref{flaws}). 


%

\item \textbf{IN with lists:} \ESM allows the keyword \IN followed by a subquery, but doesn't allow a list of values. 
\ETM properly parses and evaluates lists within the \IN keyword (Listing~\ref{lst:in_list}).

\vspace{0.5em}
\begin{lstlisting}[language=SQL,keywordstyle=\color{blue},numbers=none,framexleftmargin=0.3em,xleftmargin=.7em,xrightmargin=.3em,caption=\ESM disregards this query as it cannot parse a list of values., label=lst:in_list]
SELECT c1 FROM t1 WHERE c1 IN (1, 2, 3);
\end{lstlisting}
\vspace{0.5em}


\item \textbf{Complex Queries:} \ESM only allows for a single subquery, intersection, or union operator. 
\ETM correctly allows any query to be parsed.

\item \textbf{Retrieval from Subquery:} Queries retrieving columns from the subquery are not properly parsed by \ESM. 
An example of this is \lstinline[language=SQL,keywordstyle=\color{blue},numbers=none,frame=none,basicstyle=\normalsize\ttfamily,postbreak=]!SELECT c1 FROM (SELECT * FROM t1)!. 
\ETM properly allows retrieving columns from subqueries.

\item \textbf{Parentheses:} Queries using parentheses to order conditional statements are not handled correctly by \ESM (Listing~\ref{lst:parentheses}).
\ETM correctly handles parentheses.

\vspace{0.5em}
\begin{lstlisting}[language=SQL,keywordstyle=\color{blue},numbers=none,framexleftmargin=0.3em,xleftmargin=.7em,xrightmargin=.3em,caption=\ESM incorrectly parses this the same with and without parentheses., label=lst:parentheses]
SELECT c1 FROM t1 WHERE c1 = x AND (c2 = y OR c1 = z);
\end{lstlisting}
\vspace{0.5em}

\item \textbf{Alias definition:} In \ESM, only table names can have aliases, and they must be defined with the optional \AS keyword.
\ETM properly evaluates all aliases, including for columns and expressions, and correctly allows aliases to be defined without \AS.

\item \textbf{Quote Types:} In SQL, single quotes are treated as a literal, while double quotes can be used for column names or literals. 
\ESM incorrectly treats all quotes the same, while \ETM correctly handles different quote types.
\end{enumerate}

\lstset{
  aboveskip=-2mm,
  belowskip=-4mm,
  basicstyle=\ttfamily\small,
  numbers=none,
  frame=none,
  language=SQL
}
\begin{table}[htp!]
\centering\small
\caption{Equivalent queries with no verifiable assumption implemented in \ETM.}
\label{tab:rules_no_assumptions}

\begin{tabular}{c|>{\raggedright\arraybackslash}p{11cm}}

\textbf{ID} & \multicolumn{1}{c}{\textbf{Equivalent Queries}} \\
\hline
\multirow{2}{*}{17} &
\begin{lstlisting}[language=SQL,keywordstyle=\color{blue},numbers=none,frame=none]
SELECT _ FROM _ ORDER BY c1;
\end{lstlisting}\vspace{-0.5em} \\
& \begin{lstlisting}[language=SQL,keywordstyle=\color{blue},numbers=none,frame=none]
SELECT _ FROM _ ORDER BY JULIANDAY(c1);
\end{lstlisting} \\
\hline

\multirow{2}{*}{18} &
\begin{lstlisting}[language=SQL,keywordstyle=\color{blue},numbers=none,frame=none]
SELECT _ FROM _ WHERE c1 IN/NOT IN (x, y,...);
\end{lstlisting}\vspace{-0.5em} \\
& \begin{lstlisting}[language=SQL,keywordstyle=\color{blue},numbers=none,frame=none]
SELECT _ FROM _ WHERE c1 =/!= x OR/AND c1 =/!= y OR/AND ...;
\end{lstlisting} \\
\hline

\multirow{2}{*}{19} &
\begin{lstlisting}[language=SQL,keywordstyle=\color{blue},numbers=none,frame=none]
SELECT t1.c1 FROM t1 JOIN t2 on t1.c1 = t2.c2;
\end{lstlisting}\vspace{-0.5em} \\
& \begin{lstlisting}[language=SQL,keywordstyle=\color{blue},numbers=none,frame=none]
SELECT t2.c2 FROM t1 JOIN t2 on t1.c1 = t2.c2;
\end{lstlisting} \\
\hline

\multirow{2}{*}{20} &
\begin{lstlisting}[language=SQL,keywordstyle=\color{blue},numbers=none,frame=none]
SELECT _ FROM t1 WHERE c1 IN (SELECT c1 FROM t1 WHERE d1);
\end{lstlisting}\vspace{-0.5em} \\
& \begin{lstlisting}[language=SQL,keywordstyle=\color{blue},numbers=none,frame=none]
SELECT _ FROM t1 WHERE d1; 
\end{lstlisting} \\
\hline

\multirow{2}{*}{21} &
\begin{lstlisting}[language=SQL,keywordstyle=\color{blue},numbers=none,frame=none]
q1;
\end{lstlisting}\vspace{-0.5em} \\
& \begin{lstlisting}[language=SQL,keywordstyle=\color{blue},numbers=none,frame=none]
q1 UNION/INTERSECT q1;
\end{lstlisting} \\
\hline
\multirow{2}{*}{22} &
\begin{lstlisting}[language=SQL,keywordstyle=\color{blue},numbers=none,frame=none]
SELECT _ FROM t1 WHERE c1 BETWEEN x AND y;
\end{lstlisting}\vspace{-0.5em} \\ 
& \begin{lstlisting}[language=SQL,keywordstyle=\color{blue},numbers=none,frame=none]
SELECT _ FROM t1 WHERE c1 >= x/y and c1 <= x/y; 
\end{lstlisting} \\
\hline
\multirow{2}{*}{23} &
\begin{lstlisting}[language=SQL,keywordstyle=\color{blue},numbers=none,frame=none]
SELECT _ FROM t1 WHERE c1 !=/>/</>=/<=/= x;
\end{lstlisting}\vspace{-0.5em} \\ 
& \begin{lstlisting}[language=SQL,keywordstyle=\color{blue},numbers=none,frame=none]
SELECT _ FROM t1 WHERE NOT c1 =/<=/>=/</>/!= x; 
\end{lstlisting} \\
\hline
\multirow{2}{*}{24} &
\begin{lstlisting}[language=SQL,keywordstyle=\color{blue},numbers=none,frame=none]
SELECT CASE WHEN d1 THEN x ELSE y END;
\end{lstlisting}\vspace{-0.5em} \\ 
& \begin{lstlisting}[language=SQL,keywordstyle=\color{blue},numbers=none,frame=none]
SELECT IIF(d1, x, y);
\end{lstlisting} \\
\hline
\multirow{2}{*}{25} &
\begin{lstlisting}[language=SQL,keywordstyle=\color{blue},numbers=none,frame=none]
SELECT _ FROM t1 LEFT JOIN t2 on t1.c1 = t2.c2 WHERE t2._ IS NULL;
\end{lstlisting}\vspace{-0.5em} \\ 
& \begin{lstlisting}[language=SQL,keywordstyle=\color{blue},numbers=none,frame=none]
SELECT _ FROM t1 WHERE t1.c1 NOT IN (SELECT c2 FROM t2);
\end{lstlisting} \\
\hline
\multirow{2}{*}{26} &
\begin{lstlisting}[language=SQL,keywordstyle=\color{blue},numbers=none,frame=none]
WITH q AS (q1) SELECT _ FROM q;
\end{lstlisting}\vspace{-0.5em} \\ 
& \begin{lstlisting}[language=SQL,keywordstyle=\color{blue},numbers=none,frame=none]
SELECT _ FROM (q1);
\end{lstlisting}
\end{tabular}
\end{table}
\lstset{
    basicstyle=\footnotesize\ttfamily,
    columns=fullflexible,
    breaklines=true,
    breakatwhitespace=true,
    postbreak=\mbox{\textcolor{red}{$\hookrightarrow$}\space},
    breakindent=0pt,
    captionpos=b,
    numbers=left,  
    numberstyle=\tiny\color{gray},  
    stepnumber=1,  
    numbersep=4pt,  
    xleftmargin=1.2em,  
    xrightmargin=-1.2em,  
    frame=single,  
    framexleftmargin=1em,  
    keepspaces=true,
    aboveskip=2mm,
    belowskip=0mm,
}

\section{Results}

\label{sec:evaluation_existing}

\subsection{Spider Models}
\label{ssec:spider-models}

Three PLM-based and four FLM-based models are evaluated on the Spider dataset \cite{yu2019spider}.
Section~\ref{sec:related-work} describes these models.
Below are their names as listed on the leaderboard:\footnote{\url{https://yale-lily.github.io/spider}}

{\small
\begin{itemize}
\setlength\itemsep{0em}

\item \textbf{DAIL} (PLM): DAIL-SQL + GPT4 \cite{gao2023texttosql}\\\url{https://github.com/BeachWang/DAIL-SQL}
\item \textbf{DIN} (PLM): DIN-SQL$\,$+$\,$GPT4 \cite{pourreza2023dinsql}\\\url{https://github.com/MohammadrezaPourreza/Few-shot-NL2SQL-with-prompting}
\item \textbf{C3} (PLM): C3 + ChatGPT + Zero-Shot \cite{dong2023c3}\\\url{https://github.com/bigbigwatermalon/C3SQL}
\item \textbf{R+N} (FLM): RESDSQL-3B + NatSQL \cite{li2023resdsql}\\\url{https://github.com/RUCKBReasoning/RESDSQL}
\item \textbf{G+P} (FLM): Graphix-3B + PICARD \cite{li2023graphixt5}\\\url{https://github.com/AlibabaResearch/DAMO-ConvAI/tree/main/graphix}
\item \textbf{R+P} (FLM): RASAT + PICARD \cite{qi2022rasat}\\\url{https://github.com/LUMIA-Group/rasat}
\item \textbf{CodeS} (FLM): CodeS-7b \cite{li2024codesbuildingopensourcelanguage}\\\url{https://github.com/RUCKBReasoning/codes}
\item \textbf{Super} (FLM): SuperSQL \cite{supersql2024}\\\url{https://github.com/BugMaker-Boyan/NL2SQL360}

\end{itemize}
}

\noindent We obtain the outputs for DAIL, DIN, C3, G+P, CodeS, and Super on the development set from their repositories, and reproduce the outputs for R+N and R+P using their sources.
We reproduce the outputs for all models on the evaluation set, as most were introduced before it was released.

\subsection{BIRD Models}
\label{ssec:bird-models}

Five models are evaluated on the BIRD dataset \cite{BIRD2023}:

{\small
\begin{itemize}
\setlength\itemsep{0em}

\item \textbf{DAIL}: the same model as described in Section~\ref{ssec:spider-models}
\item \textbf{C3}: the same model as described in Section~\ref{ssec:spider-models}
\item \textbf{RESD}: R+N without NatSQL \cite{li2023resdsql}
\item \textbf{CodeS-15}: 15b version of CodeS
\item \textbf{Super}: the same model as described in Section~\ref{ssec:spider-models}

\end{itemize}
}

\subsection{Results}
\label{ssec:results}

Table~\ref{tab:results_spider} shows the results of the seven Spider models (Section~\ref{ssec:spider-models}) with respect to \EXE, \ESM, and \ETM.
For the development set, Super performs the best on \EXE, while R+N performs best on \ESM. 
Super's \EXE score is 2.6\% higher than the next best model, CodeS, although its \ESM score is 7.3\% lower.
This discrepancy is diminished to 1.5\% with \ETM; however, this still indicates it is not as strong as CodeS overall.
DAIL, despite scoring better than every FLM model except CodeS on \EXE, scores significantly worse on \ETM, getting outperformed by both R+N and G+P.
The trend is evident; FLM-based models exhibit 3-7\% decreases in performance from \ESM to \ETM, whereas PLM-based models show 1-14\% increases.
This impact is especially dramatic for zero-shot methods.
For example, C3 performs relatively well on \EXE (only 3.7\% lower than CodeS) but extremely poorly on \ESM (32.5\% lower than CodeS), and then substantially recovers on \ETM (16.2\% lower than CodeS).
Likewise, DIN also underperforms on \ESM and thus gets a significant 4.6\% boost when evaluated using \ETM.
We attribute this to  \ESM's limitations in handling query styles that deviate from the Spider dataset.
FLM-based models are less impacted because they are trained to learn those styles, whereas PLM-based models--which often generate queries with styles not captured in the training set--are often semantically correct but still get penalized by \ESM.

\begin{table}[htp!]
\centering\small
\caption{Model performance on the Spider dataset in \%. Column-wise rankings are indicated in parentheses. The \textbf{Evaluation Set} columns display the results from the model outputs reproduced by us, while the \textbf{Reported} columns show the results on the evaluation set as reported in the respective literature and the leaderboard for those models.}
\label{tab:results_spider}

\begin{tabular}{cc|ccc|ccc|cc}
\toprule
\multicolumn{2}{c|}{\multirow{2}{*}{\bf Model}} & \multicolumn{3}{c|}{\bf Development Set} & \multicolumn{3}{c|}{\bf Evaluation Set} & \multicolumn{2}{c}{\bf Reported} \\
 & & \textbf{\EXE} & \textbf{\ESM} & \textbf{\ETM} & \textbf{\EXE} & \textbf{\ESM} & \textbf{\ETM} & \textbf{\EXE} & \textbf{\ESM} \\
\midrule
\bf DAIL & PLM &         82.9 (3) &     70.0 (6) &     71.5 (5) &     82.2 (3) &     66.1 (4) &     68.1 (5) &     86.2 (2) &     66.5 (4) \\
\bf DIN  & PLM &         81.7 (5) &     60.1 (7) &     64.7 (7) &     81.6 (4) &     60.7 (6) &     64.8 (6) &     85.3 (3) &     60.0 (5) \\
\bf C3   & PLM &         79.8 (7) &     46.9 (8) &     59.8 (8) &     79.5 (5) &     43.9 (7) &     58.5 (7) &     82.3 (4) &     -        \\
\bf Super & PLM &    \bf 86.1 (1) &     72.1 (5) &     75.1 (2) & \bf 85.3 (1) &     65.5 (5) &     70.4 (2) & \bf 87.0 (1) &     -        \\
\midrule
\bf R+N  & FLM &         82.8 (4) & \bf 80.5 (1) &     74.9 (3) &     78.4 (6) &     70.9 (2) &     70.4 (2) &     79.9 (5) &     72.0 (2) \\
\bf G+P  & FLM &         80.1 (6) &     77.1 (3) &     72.3 (4) &     -        &     -        &     -        &     77.6 (6) & \bf 74.0 (1) \\
\bf R+P  & FLM &         76.7 (8) &     75.2 (4) &     69.3 (6) &     77.9 (7) &     69.3 (3) &     69.6 (4) &     75.5 (7) &     70.9 (3) \\
\bf CodeS & FLM &        83.5 (2) &     79.4 (2) & \bf 76.6 (1) &     83.0 (2) & \bf 73.7 (1) & \bf 73.2 (1) &     -        &     -        \\

\bottomrule
\end{tabular}
\end{table}

On the evaluation set, the trend between \ESM and \ETM remains consistent.
Overall, PLM-based models dominate FLM-based models on \EXE: the best PLM-based model, Super, gives a 2.3\% higher score than the best FLM-based model, CodeS.\footnote{Unfortunately, we were unable to run the G+P model, so its results on the evaluation set are omitted from Table~\ref{tab:results_spider}.}
This pattern continues with the other PLM models, which all outperform R+N and R+P on \EXE.
However, for \ESM, CodeS outperforms Super by 8.2\%, although the gap narrows to 2.8\% on \ETM.
Note that the \EXE scores of PLM-based models decrease from their originally reported values to our replicated results, while FLM-based models see similar or slightly higher scores.\footnote{CodeS's scores on the evaluation set weren't reported, and CodeS was not submitted to the Spider leaderboard.}
We attribute these differences to the absence of distinct and value checking (Section~\ref{ssec:sql-equivalence}), as well as the high variance in PLM-based approaches, discussed further in Section~\ref{sec:plm_variance}.

Table~\ref{tab:results_bird} illustrates the results of the five models evaluated on BIRD's development set (Section~\ref{ssec:bird-models}).
All models perform poorly on \ESM, largely because the parser in the evaluation script fails to handle the complexity of queries in the BIRD dataset.
Consequently, run-time errors are common, and the script automatically classifies queries with such errors as syntactically incorrect, leading to notably low scores and indicating that \ESM is not a suitable metric for a model's performance on this dataset.
By contrast, \ETM--unaffected by these shortcomings--still shows a marked difference from \EXE. 
Similarly to the Spider results, models across the board perform worse on \ETM than on \EXE, though some are affected more severely than others.
While Super achieves the highest \EXE score, it is surpassed by CodeS-15 on \ETM.
C3 and RESD have a similar trend; despite outperforming RESD on \EXE, C3 suffers a bigger drop off from \EXE to \ETM, causing RESD to score higher on \ETM.

\begin{table}[htbp!]
\centering\small
\caption{Model performance (\%) on the BIRD development set. Rankings are indicated in parentheses.}
\label{tab:results_bird}
\begin{tabular}{cc|ccc}
\toprule
\multicolumn{2}{c|}{\multirow{2}{*}{\bf Model}} & \multicolumn{3}{c}{\bf Development Set}\\
 & & \textbf{\EXE} & \textbf{\ESM} & \textbf{\ETM}  \\
\midrule
\bf DAIL      & PLM    & 50.1 (3) & 8.0 (3) & 31.9 (3) \\
\bf C3        & PLM    & 42.8 (4) & 5.7 (5) & 22.6 (5) \\
\bf Super     & PLM & \bf 52.1 (1) & 8.3 (2) & 33.1 (2) \\
\midrule
\bf RESD      & FLM    & 37.4 (5) & 7.2 (4) & 25.4 (4) \\
\bf CodeS-15  & FLM    & 51.6 (2) & \bf 9.1 (1) & \bf 36.2 (1) \\
\bottomrule
\end{tabular}
\end{table}

\noindent A notable shift occurs in model rankings from \EXE to \ETM, as PLM-based models that dominate on \EXE often underperform relative to FLM-based models on \ETM.
A likely explanation is that PLMs, which excel in broader language understanding, make implicit assumptions about the databases that prove problematic when the queries are evaluated on \ETM.
As discussed further in Section~\ref{sec:evaluation}, these types of assumptions typically do not affect performance except in edge cases which may be absent in the dataset.
In contrast, FLMs--though weaker on \EXE--may avoid some of these pitfalls, resulting in higher \ETM scores.
Meanwhile, the performance decrease from \ESM to \ETM for the FLM-based models on Spider likely stems from their lack of optimization for certain SQL features that \ESM does not assess (Section~\ref{sec:revision}), causing incorrect handling of those (e.g., generating random conditions for \JOIN would not have any impact on \ESM).
These findings highlight the need for more robust evaluation metrics, such as \ETM, to facilitate further improvements in Text-to-SQL.

\section{Discussion}

\label{sec:analysis}

\subsection{Model Evaluation}
\label{sec:evaluation}

Upon analysis of why PLM-based models achieve high \EXE scores but score lower on \ETM, we find that they often generate queries that are equivalent to the gold query under certain table-specific verifiable assumptions.
However, these assumptions are not enforced by the actual schema of those tables, so \ETM correctly identifies that the queries are semantically distinct.
In practice, the assumptions may hold for the tables in the database at hand, leading to false positives in \EXE, since the predicted queries are not guaranteed to produce correct results in other databases.
In such scenarios, \ETM is a more robust metric than \EXE, as it yields fewer false positives.

For Spider, C3 and DIN produce much lower \ESM scores compared to other models.
Although DIN and C3 employ highly specialized prompts taylored to the dataset's style--such as calibration hints and elaborate classification prompting--the other models make more extensive use of the training set.
The FLM-based models, for instance, are directly finetuned on the training set, thereby imitating its query style.
DAIL and Super both search the training set for questions similar to the input and use them for few-shot prompting.
Because DIN and C3 do not directly leverage the training set, they exhibit greater creativity in query generation, precisely the behavior penalized by \ESM.
This penalty is alleviated with \ETM, where the performance gaps shrink and more accurately represent model effectiveness.

The same reasoning explains why FLM-based models do not improve from \ESM to \ETM in Spider.
Their generation styles largely match the dataset's style, and \ESM already accounts for the primary assumptions relevant to that style, so only a few of the verifiable equivalence rules introduced in \ETM apply.
Moreover, certain issues present in \ESM that are addressed in our new metric (Section~\ref{ssec:etm}) cause some outputs to be evaluated more rigorously, lowering their \ETM scores relative to \ESM.

\subsection{PLM Variance}
\label{sec:plm_variance}

The discrepancy between the published results and our reproduced results on \EXE for the PLM-based models in Table~\ref{tab:results_spider} is in part due to the high natural variability inherent in PLMs.
This variability not only hinders the replicability of the work but also creates a situation where, given enough attempts, even a worse model can outperform a more consistent model.
This is exacerbated when \EXE is used as the primary evaluation metric, since many of the tables in BIRD and especially Spider do not have sufficient edge cases to catch all the false assumptions made by these models.

\ETM, however, aims to reduce the variance in PLMs by being more stringent, so that it forces the models to generate a query that \textit{always} predicts the correct values, which is much more challenging, but leads to less variance in results when evaluated under \ETM.

\subsection{Error Analysis}

To assess whether our new metric gives a more accurate evaluation, we analyzed the false positives and false negatives generated by \EXE, \ESM, and \ETM for each model on the Spider evaluation set and the BIRD development set.
Since disabling either distinct or value checks both lead to numerous false positives in both \EXE and \ESM, and most current state-of-the-art models predict values, we enabled these checks for our analysis.

Table~\ref{FalsePosNeg} presents the error analysis results. 
Despite enabling distinct and value checks, \EXE and \ESM still yield a high volume of false positives and false negatives, respectively. 
Notably, \ESM's low false positive rate on BIRD partially stems from poor internal parsing, as it very rarely evaluates any two queries in BIRD as equivalent.
For all models, the amount of false positives from \EXE and false negatives from \ESM decreases significantly in \ETM. 
The decrease in false positives from \EXE stems from the new constraints in \ETM that correctly identify mismatches, while the false negative decrease from \ESM is attributed to our equivalence rules in Table~\ref{tab:rules_assumptions}.
The false positive rate in \ESM results from the issues described in Section~\ref{fixes}, which are fixed in \ETM.

\begin{table}[htbp!]
\centering\small
\caption{False positives and negative rates (\%) for all models with respect to the three metrics on the Spider evaluation set and BIRD development set. \ESM and \EXE are evaluated with distinct and value checking enabled.}
\label{FalsePosNeg}
\begin{tabular}{c|cc|cc|cc|cc}
\midrule
\multicolumn{3}{c|}{\multirow{2}{*}{\bf Model}} & \multicolumn{2}{c|}{\textbf{\EXE}}& \multicolumn{2}{c|}{\textbf{\ESM}} & \multicolumn{2}{c}{\textbf{\ETM}} \\
\multicolumn{3}{c|}{} &  \textbf{FP} & \textbf{FN} & \textbf{FP} & \textbf{FN} & \textbf{FP} & \textbf{FN} \\
\midrule
\multirow{7}{*}{\rotatebox{90}{Spider}} & \textbf{DAIL} & P  &  16.3 & 0.0 & 5.0 & $\:\:$3.2 & 0.1 & 0.0 \\
& \textbf{DIN} & P   & 19.5 & 0.0 & 6.1 & $\:\:$9.8 & 0.0 & 0.0  \\
& \textbf{C3} & P    & 23.0 & 0.0 & 2.8 &      17.5 & 0.1 & 0.0  \\
& \textbf{Super} & P    & 15.0 & 0.0 & 3.2 &  $\:\:$8.2 & 0.0 & 0.0  \\
& \textbf{R+N} & F   & 10.0 & 0.0 & 4.6 & $\:\:$3.8 & 0.1 & 0.0  \\
& \textbf{R+P} & F   & 10.0 & 0.0 & 4.1 & $\:\:$4.3 & 0.1 & 0.0  \\
& \textbf{CodeS} & F & 12.1 & 0.0 & 4.7 & $\:\:$4.0 & 0.0 & 0.0  \\
\midrule
\multirow{6}{*}{\rotatebox{90}{BIRD}} & \textbf{DAIL}   & P &  17.4 & 0.0 & 1.6 & 26.3 & 0.3 & 1.1 \\
& \textbf{C3}    & P & 17.5 & 0.0 & 1.1 & 20.7 & 0.0 & 2.7 \\
& \textbf{Super} & P & 17.9 & 0.0 & 1.2 & 27.1 & 0.2 & 1.2 \\
& \textbf{RESD}  & F & 11.1 & 0.0 & 1.7 & 20.7 & 0.1 & 1.0 \\
& \textbf{CodeS} & F & 14.8 & 0.0 & 1.2 & 28.9 & 0.1 & 0.8 \\
\end{tabular}
\end{table}

\noindent On both \EXE and \ESM, false positives and false negatives disproportionately affect certain models.
In particular, PLM-based models have higher false positive rates for \EXE and higher false negative rates on \ESM than FLM-based ones.
In contrast, \ETM exhibits a notably smaller discrepancy between the best and worst models, indicating it is less biased than either \EXE or \ESM.
The few remaining false positives stem from inconsistencies in the database schema itself, where the actual database does not comply with the specified schema requirements.
On Spider, the false negative rate is entirely eliminated, whereas on BIRD, a small number of false negatives persist.
Although this is far fewer than the false positives for \EXE, it suggests that additional equivalence rules could further reduce the false negative rate, which we will explore in future work.

\subsection{Equivalence Rule Analysis}
\begin{figure}[htbp!]
\centering
\includegraphics[width=0.8\textwidth]{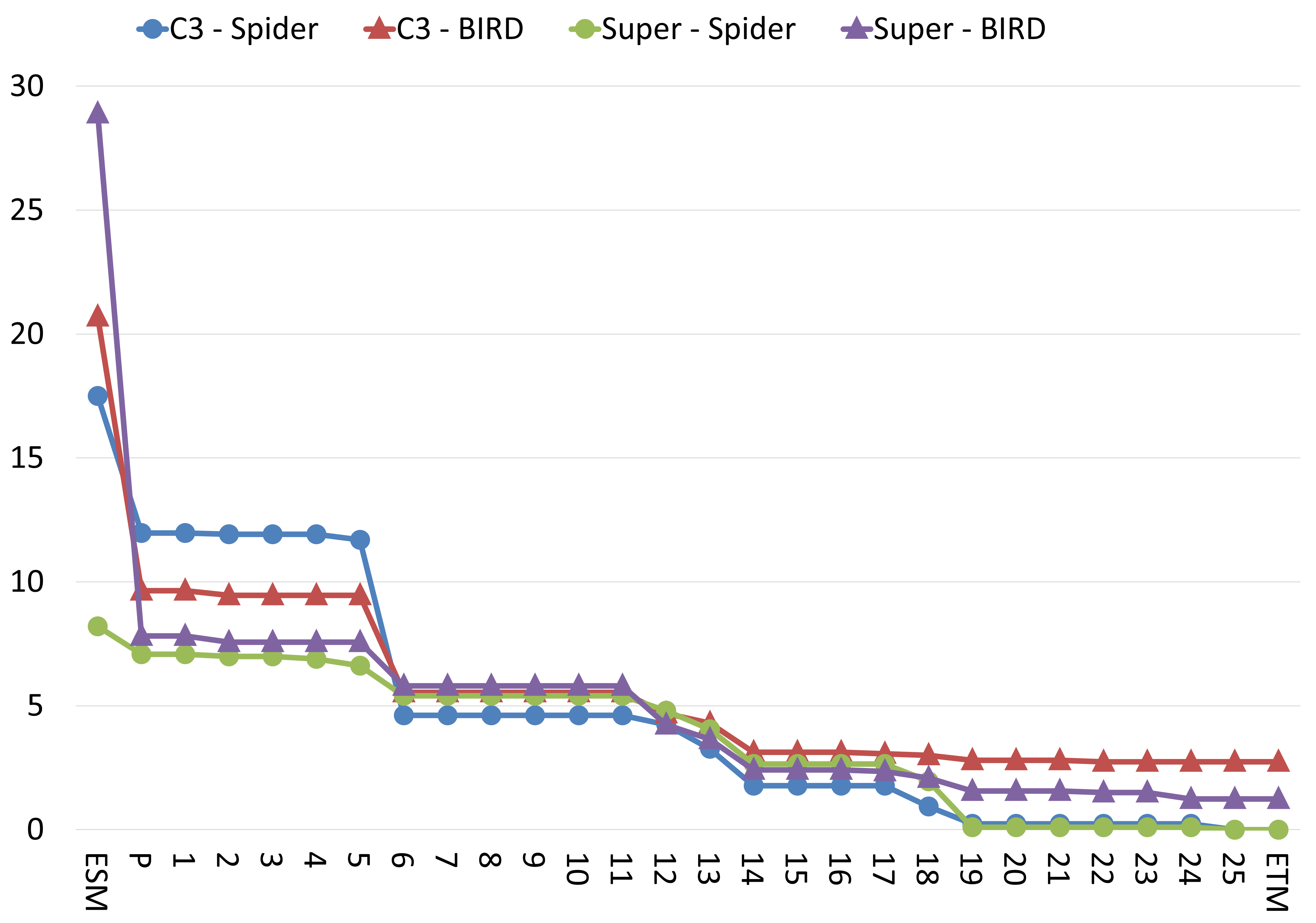}
\caption{False negative rates of \ETM (\%) on selected models for Spider and BIRD as our equivalence rules are accumulated. \textbf{\ESM}: the original \ESM metric, \textbf{P}: \ETM using only the preprocessing equivalence rules (Table~\ref{tab:rules_assumptions}), \bm{$n$}: \textbf{P} + equivalence rules 1 to $n$ in Table~\ref{tab:rules_assumptions} are applied, \textbf{\ETM}: \textbf{P} + all 26 equivalence rules are applied, which is our final \ETM. Full results can be found in Appendix~\ref{ssec:FN_full}.}
\label{fig:rules_negatives}
\end{figure}

To look into this further, we perform an analysis of improvement of the \ETM metric as equivalence rules are added (Figure~\ref{fig:rules_negatives}). 
When only the preprocessing equivalence rules are used, there is already a large decrease in false negative rate from \ESM to \ETM in both models on BIRD and the C3 model on Spider due to the fixes in functionality (Section~\ref{ssec:etm}), while the Super model on Spider had a much smaller decrease, because it was imitating Spider's style, which \ESM was built for.
Without these preprocessing rules, the basic AST matching has much higher false negative rates (shown in Appendix~\ref{ssec:FN_full}).

The overall trend shows that as expected, each rule we cumulatively add decreases the false negative rate of \ETM.
However, the equivalence rules did not have equal impact on the improvement from \ESM to \ETM.
Some models benefitted more from certain rules, and others didn't have a reliance on any one rule in particular.
Some rules were particularly impactful, while others were almost never utilized.
Rule 14 (which entailed unecessary use of \JOIN) was the most important addition in Spider for models that directly used the training set (DAIL, Super, FLM-based) because even the training sets were not consistent in whether the \JOIN keywords they used were necessary. 
Thus, the models that relied on it had similar levels of variation. 
The models that were more reliant on generating without access to examples from the training set, however, had much more reliance on specific rules, indicating a certain preference for styles of SQL queries.
For example, the models based solely on GPT (C3 and DIN) had Rule 6 as the most useful, indicating that GPT has a bias towards generating \COUNT(c1) instead of \COUNT(*).
As more rules are added, the discrepancy in false negatives between the best and worst models decreases, showing that each equivalence rule added reduces bias in \ETM.

\section{Conclusions}
\label{sec:conclusion}
%
This study introduces Enhanced Tree Matching (\ETM), a novel evaluation metric for Text-to-SQL that overcomes several limitations of the previous metrics, Execution (\EXE) and Exact Set Matching (\ESM).
Our findings indicate that \ETM offers a substantial improvement by reducing the occurrences of both false positives and false negatives that commonly plague the earlier metrics. 
By adopting a more rigorous approach and incorporating verifiable equivalence rules to allow query diversity, \ETM can discern more granular distinction in query correctness, allowing for a more accurate measurement of the semantic accuracy of the generated queries and a better understanding of LLMs' true capabilities in generating SQL queries. 
While \EXE may be suitable for applications with few edge cases, domains where errors must be avoided demand a more rigorous metric like \ETM.

Moving forward, we plan to extend the list of verifiable rules to strengthen \ETM with the help of community feedback, thereby increasing its robustness in evaluating complex SQL query structures. With our framework, adding new rules to strengthen our metric is much easier than improving \ESM.
As we continue to refine and enhance \ETM, our goal is to establish a new standard for evaluating Text-to-SQL models that can accurately represent their practical utility and technical proficiency in real-world applications. 
With the introduction of \ETM, we hope that more PLM-based methods will tackle datasets like CoSQL, which is currently evaluated with \ESM, since they will no longer be constrained by the lack of variation enforced by \ESM.
\vspace{6pt} 



\authorcontributions{Conceptualization, B.A and J.C; methodology, B.A and J.C; formal analysis, B.A and Y.K; investigation, B.A, software, B.A and Y.K, writing---original draft prepareation, B.A and J.C., writing---review and editing, B.A and J.C, supervision, J.C, funding acquisition, J.C. All authors have read and agreed to the published version of the manuscript.}

\funding{This research received no external funding.}

\institutionalreview{Not applicable.}

\dataavailability{Our released evaluation script is publicly available
under the Apache 2.0 license on \url{https://github.com/emorynlp/ETM/}.}

\acknowledgments{We greatly acknowledge the support of Emory University.}

\conflictsofinterest{The authors declare no conflict of interest.} 

\appendixtitles{yes} 
\appendixstart
\appendix
\section[\appendixname~\thesection]{}
\subsection{False Negatives with Equivalent Rules}
\label{ssec:FN_full}

\begin{figure}[htbp!]
  \centering
  \begin{subfigure}{\textwidth}    
    \centering
    \includegraphics[width=\textwidth]{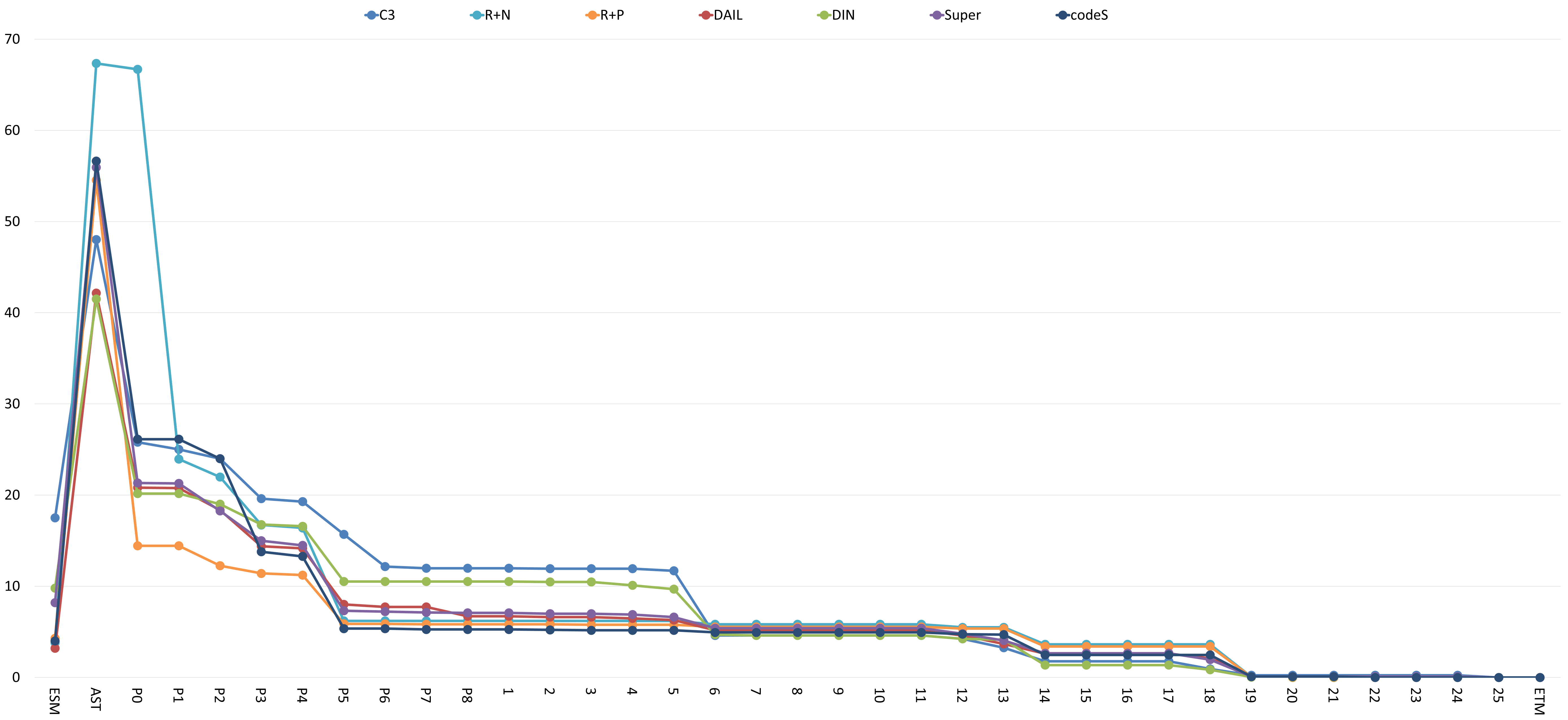}
    \caption{False negative rates on spider.}
    \label{fig:FN_full_spider}
  \end{subfigure}
  
  \vspace{1em}
  
  \begin{subfigure}{\textwidth}    
    \centering
    \includegraphics[width=\textwidth]{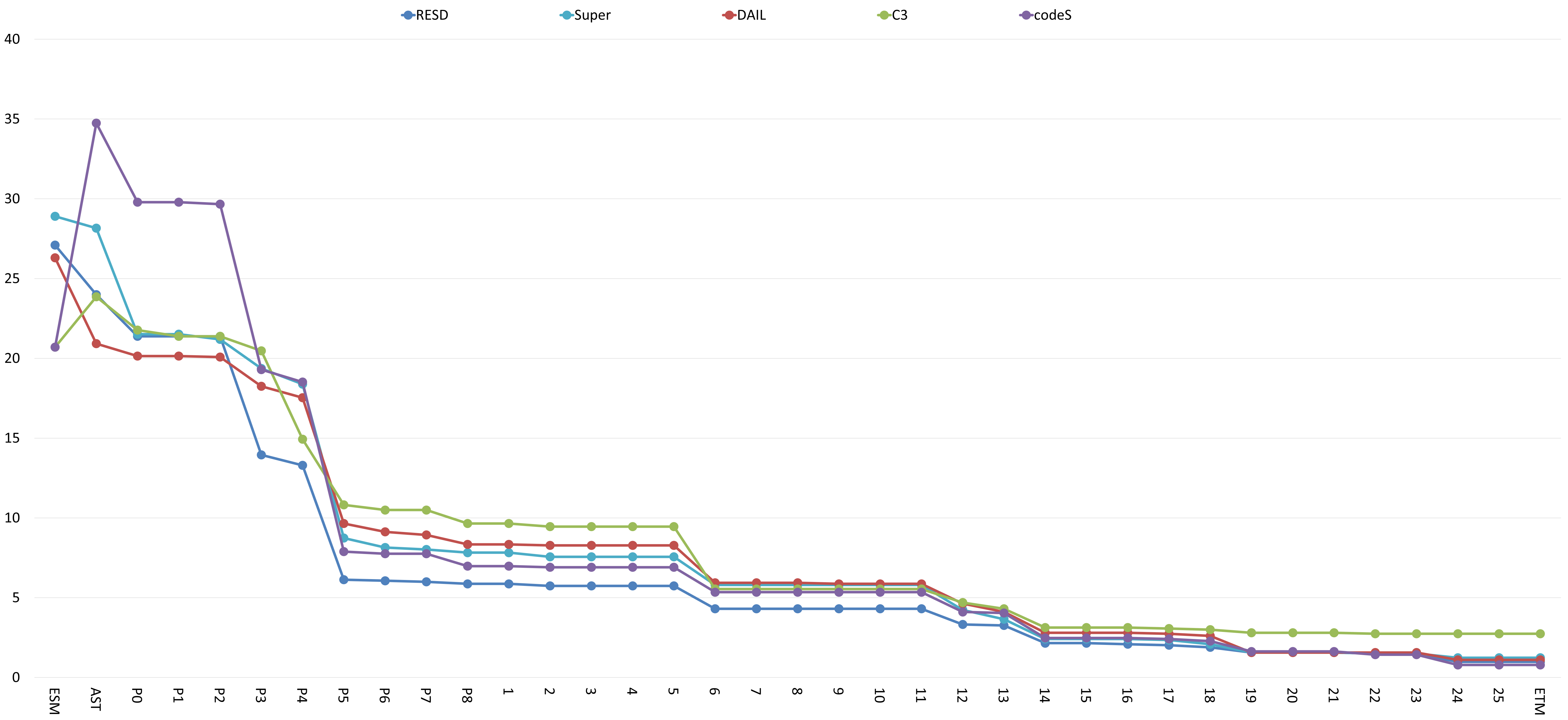}
    \caption{False negative rates on BIRD.}
    \label{fig:FN_full_bird}
  \end{subfigure}
  
  \caption{False negative rates for \ETM (\%) on spider (\ref{fig:FN_full_spider}) and BIRD (\ref{fig:FN_full_bird}) as our equivalence rules are accumulated. \textbf{\ESM}: the original \ESM metric, \textbf{P}\bm{$n$}: \ETM using rules $P0$ to $Pn$ (Table~\ref{tab:rules_assumptions}), \bm{$n$}: \textbf{P8} + equivalence rules 1 to $n$ in Table~\ref{tab:rules_assumptions} are applied, \textbf{\ETM}: \textbf{P} + all 26 equivalence rules are applied, which is our final \ETM.}
  \label{fig:FN_full}
\end{figure}

\begin{adjustwidth}{-\extralength}{0cm}

\clearpage
\reftitle{References}


\bibliography{custom}

\end{adjustwidth}
\end{document}